\documentclass[runningheads]{llncs}
\usepackage{graphicx}
\pdfpkmode{ljfzzz}
\usepackage{cite}
\usepackage[T4,T1]{fontenc}
\usepackage{hyperref}
\usepackage{multirow}
\titlerunning{Suggestive Annotation with Gradient-guided Sampling}
\usepackage{floatrow}
\usepackage{textcomp}
\newfloatcommand{capbtabbox}{table}[][\FBwidth]

\makeatletter
\def\easytfoursymbol#1{\mathord{\mathchoice
  {\mbox{\fontsize\tf@size\z@\usefont{T4}{\rmdefault}{m}{it}\char#1}}
  {\mbox{\fontsize\tf@size\z@\usefont{T4}{\rmdefault}{m}{it}\char#1}}
  {\mbox{\fontsize\sf@size\z@\usefont{T4}{\rmdefault}{m}{it}\char#1}}
  {\mbox{\fontsize\ssf@size\z@\usefont{T4}{\rmdefault}{m}{it}\char#1}}
}}
\makeatother
\newcommand{\hookD}{\easytfoursymbol{129}}
\begin{document}
\title{Suggestive Annotation of Brain Tumour Images with Gradient-guided Sampling}
%
%
\author{Chengliang Dai \inst{1} \and
Shuo Wang \inst{1} \and
Yuanhan Mo \inst{1} \and
Kaichen Zhou\inst{4} \and 
Elsa Angelini \inst{2} \and
Yike Guo \inst{1} \and
Wenjia Bai \inst{1,3}}
\authorrunning{C. Dai, et al.}

\institute{Data Science Institute, Imperial College London, London, UK
\and
ITMAT Data Science Group, Imperial College London, London, UK
\and
Department of Brain Sciences, Imperial College London, London, UK
\and
Department of Computer Science, University of Oxford
}

\maketitle              

\begin{abstract}
Machine learning has been widely adopted for medical image analysis in recent years given its promising performance in image segmentation and classification tasks. As a data-driven science, the success of machine learning, in particular supervised learning, largely depends on the availability of manually annotated datasets. For medical imaging applications, such annotated datasets are not easy to acquire. It takes a substantial amount of time and resource to curate an annotated medical image set. In this paper, we propose an efficient annotation framework for brain tumour images that is able to suggest informative sample images for human experts to annotate. Our experiments show that training a segmentation model with only 19\% suggestively annotated patient scans from BraTS 2019 dataset can achieve a comparable performance to training a model on the full dataset for whole tumour segmentation task. It demonstrates a promising way to save manual annotation cost and improve data efficiency in medical imaging applications.

\end{abstract}
\section{Introduction}
Machine learning techniques have been widely used in medical imaging applications, achieving performance comparable to or even surpassing human-level performance in a variety of tasks \cite{liang2019evaluation}. The prevalence of machine learning techniques comes with some evident challenges and one of them is the requirement for large amount of annotated data for training. In medical imaging applications, the availability of imaging data itself is usually not the real challenge since thousands of people are examined by MRI, CT or X-ray scanners everyday, generating large quantities of medical images. The real challenge comes from the scarcity of human expert annotations for these images. Annotating medical images can be expensive in terms of both time and expertise. For instance, annotating the brain tumour image of one patient may take several hours even for a skilled image analyst \cite{fiez2000lesion}.

Many efforts have been devoted to addressing the challenge of the lack of annotated data. One perspective is to maximise the use of existing data annotations. For instance, transfer learning \cite{tajbakhsh2016convolutional} and domain adaptation \cite{kamnitsas2017unsupervised} were proposed to utilise features learnt from annotations in an existing domain for applications in a target domain where annotations are limited. Semi-supervised and weakly-supervised learning \cite{cheplygina2019not} is another example, which leverages unlabelled or weakly-labelled data, together with labelled data, to achieve a better performance than using labelled data alone.

Another perspective to address the challenge is to optimise the annotation process, which is in line with what we have explored in this paper. As has been observed in \cite{katharopoulos2018not, fan2017learning}, different data samples contribute to the training of a machine learning model to different extents. Selecting more \textit{informative} samples for manual annotation and for model training can therefore potentially reduce the annotation cost and improve efficiency. To improve the efficiency of annotation, active learning \cite{settles2009active} was proposed for querying annotations for the most informative samples only. The informativeness of an unannotated data sample is normally defined by two equally important metrics: uncertainty, the confidence of the model in predicting the correct label; and representativeness, which describes how much characteristics of a data sample can represent the dataset it belongs to. To select the most informative biomedical images for annotation, multiple fully convolutional networks (FCNs) were used in \cite{yang2017suggestive} to measure the uncertainty of data and the cosine similarity was used to find representative samples among all candidates. In \cite{sharma2019active}, a coreset-based method, which used a small set of data points to approximate all of the data points, were combined with uncertainty sampling for selecting brain lesion images that were both uncertain and representative. \cite{shi2019active} proposed a dual-criteria method, which utilised local sensitive hashing and prediction confidence ranking to select informative samples. These methods have shown effectiveness in selecting samples that contribute more to model training.

In this work, we propose a gradient-guided sampling method that utilises the gradient from the training loss to guide the selection of informative samples from a data manifold learnt by a variational autoencoder (VAE) \cite{kingma2013auto}. There are two major contributions of this work: (1) Medical images are suggested for annotation using the loss gradient projected onto the data manifold. Although the loss gradient was explored in the context of adversarial attack and learning \cite{kurakin2016adversarial}, its role in suggestive annotation has not been investigated in depth. (2) To further reduce annotation cost, we takes the data redundancy in 3D volume into account, investigating and comparing patient-wise vs. image-wise suggestions. The proposed method was evaluated on a challenging annotation task, brain tumour annotation, and it achieved a promising performance.

\section{Methods}
\subsection{Overview}
\begin{figure}[]
\makebox[\textwidth][c]{\includegraphics[width=0.9\textwidth]{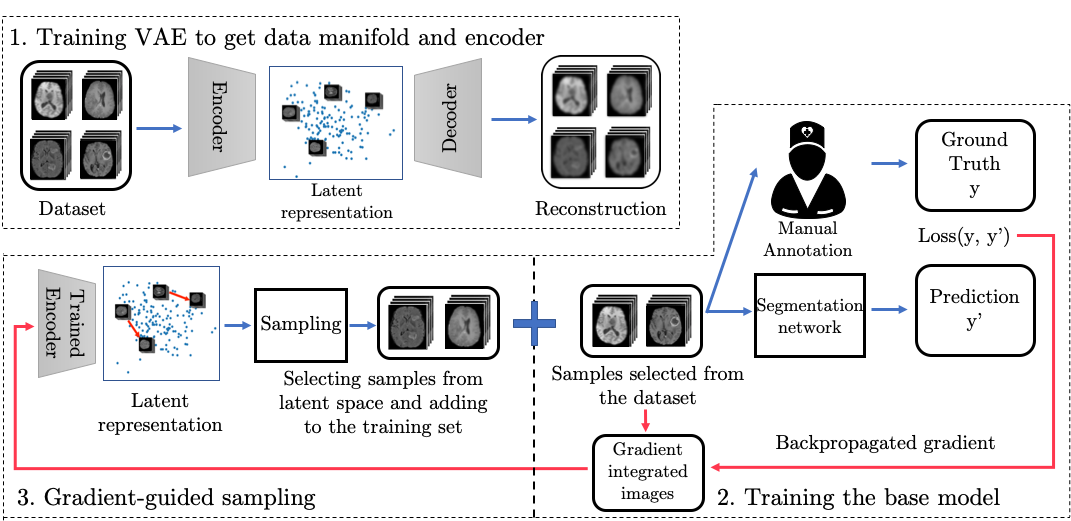}}
\caption{An overview of the suggestive annotation method. The framework consists of three steps: (1) Training a VAE for learning the data manifold. (2) Training the segmentation model and backpropagating the loss gradient to image space. (3) Sampling on the learnt manifold guided by the gradient.} \label{framework}
\end{figure}

The framework of gradient-guided suggestive annotation is illustrated in Fig. \ref{framework}, consisting of three steps: (1) A VAE is trained with an unannotated dataset to learn a data manifold in the latent space. (2) An image segmentation model (base model), where we use a 2D UNet \cite{ronneberger2015u} as an example, is trained in an active learning manner. To initialise the model, the first batch of training data is randomly selected from the unannotated dataset and suggested for annotation. After training for several epochs, the gradient of the training loss is backpropogated to the image space. (3) Gradient-integrated image is further projected to the latent space using the VAE encoder. New unannotated samples are selected in the latent space based on the gradient and suggested for the next round of annotation. The model is then trained on images using annotations from both step (2) and (3).

Step (2) and (3) can be done iteratively to suggest more batches of samples. In this work, we perform it for just one iteration as the research focus here is on investigating of whether such sampling method works and how well it works compared to other methods, such as random suggestion or oracle method.

\subsection{Learning the data manifold}
Representativeness is important for selecting informative samples. Learning a data manifold that can reflect the structure of a given dataset is the prerequisite for the proposed framework. The VAE is used here for learning the data manifold given its good potential shown in many other works \cite{liu2017unsupervised}. Let us denote an unannotated medical images dataset by \(X=\{x_1,x_2,...,x_n\}\). A VAE is trained on \(X\) with the loss function formulated as
\begin{equation}\label{eq:vaeloss}
\mathcal{L}_{vae}(\theta,\phi;x_{i}) = MSE(f_{\theta}(g_{\phi}(x_{i})),x_{i}) + D_{KL}(q_\phi(z|x_{i})||p_\theta(z)),
\end{equation}

\noindent where \(g_{\phi}(\cdot)\) and \(f_{\theta}(\cdot)\) denote the encoder and decoder, which typically consist of a number of convolutional layers \cite{kingma2013auto}. MSE denotes the mean square error function and \(D_{KL}\) denotes the KL-divergence, which regularises the optimisation problem by minimising the distance between the latent variable distribution and a Gaussian distribution \cite{kingma2013auto}. Once trained, the VAE can be used to obtain the latent representations \(Z=\{z_1,z_2,...,z_n\}\) given \(X\). It will be used for sampling purpose in the later stage.

\subsection{Training the base model}

We train the base segmentation model which can provide the gradient for sample suggestion. As revealed in \cite{kumar2010self, bengio2009curriculum}, less \textit{informative} samples are more important at the early stage of model training, while \textit{harder} samples are more important at the later stage. Therefore, we first randomly select a subset \(S=\{x_{1},x_{2},...,x_{m}\}\) of $m$ samples, which are used to initialise the model training and explore the hard samples in the manifold.

An annotated dataset \(\hookD=\{(x_{1},y_{1}),(x_{2},y_{2}),...,(x_{m},y_{m})\}\) is thus constructed, where $y$ denotes the annotation by the expert. A base model is trained on \(\hookD\) with the Dice loss function defined by,
\begin{equation}
\mathcal{L}_{Dice}(y_{i},\hat{y}_{i})= -\frac{2\sum \hat{y}_{i}y_{i}}{\sum \hat{y}_{i}+\sum y_{i}},  (x_{i},y_{i})\in \hookD.
\end{equation}
where \(\hat{y}_{i}\) denotes the output of the base model given \(x_{i}\), which is a probability map of segmentation and \(y_{i}\) denotes the ground truth annotation from the expert.

After the base model is trained, for each of the $m$ samples \(x_{i}\in S \), its gradient of loss is backpropagated to the image space according to,
\begin{equation}\label{eq:integration}
    x'_{i}=x_{i}+\alpha\frac{\partial L_{Dice}}{\partial x_{i}},
\end{equation}
\noindent where the gradient informs the direction to harder samples and \(\alpha\) denotes the step length along the gradient. 

\subsection{Gradient-guided sampling}
Here, we describe gradient-guided sampling, which selects informative samples on the learnt data manifold and strikes a balance between exploring uncertain samples and representative samples. Using the VAE encoder \(g_{\phi}(\cdot)\), the hard sample \(x'_{i}\) can be projected to the latent space by,
\begin{equation}\label{eq:projection}
    z'_{i}=g_{\phi}(x'_{i}).
\end{equation}

Theoretically, images can be synthesised from $z'_{i}$ via the VAE decoder \(f_{\theta}(\cdot)\) and suggested to the expert for annotation. However, the synthetic image may not be of high quality, which would prevent the expert from producing reliable annotation. To mitigate this issue, we propose to sample in the real image space, searching for existing real images that are most similar to the synthesised image in the latent space. In this way, the expert would be able to annotate on high-quality real images.

\begin{figure}[ht]
\makebox[\textwidth][c]{\includegraphics[width=0.8\textwidth]{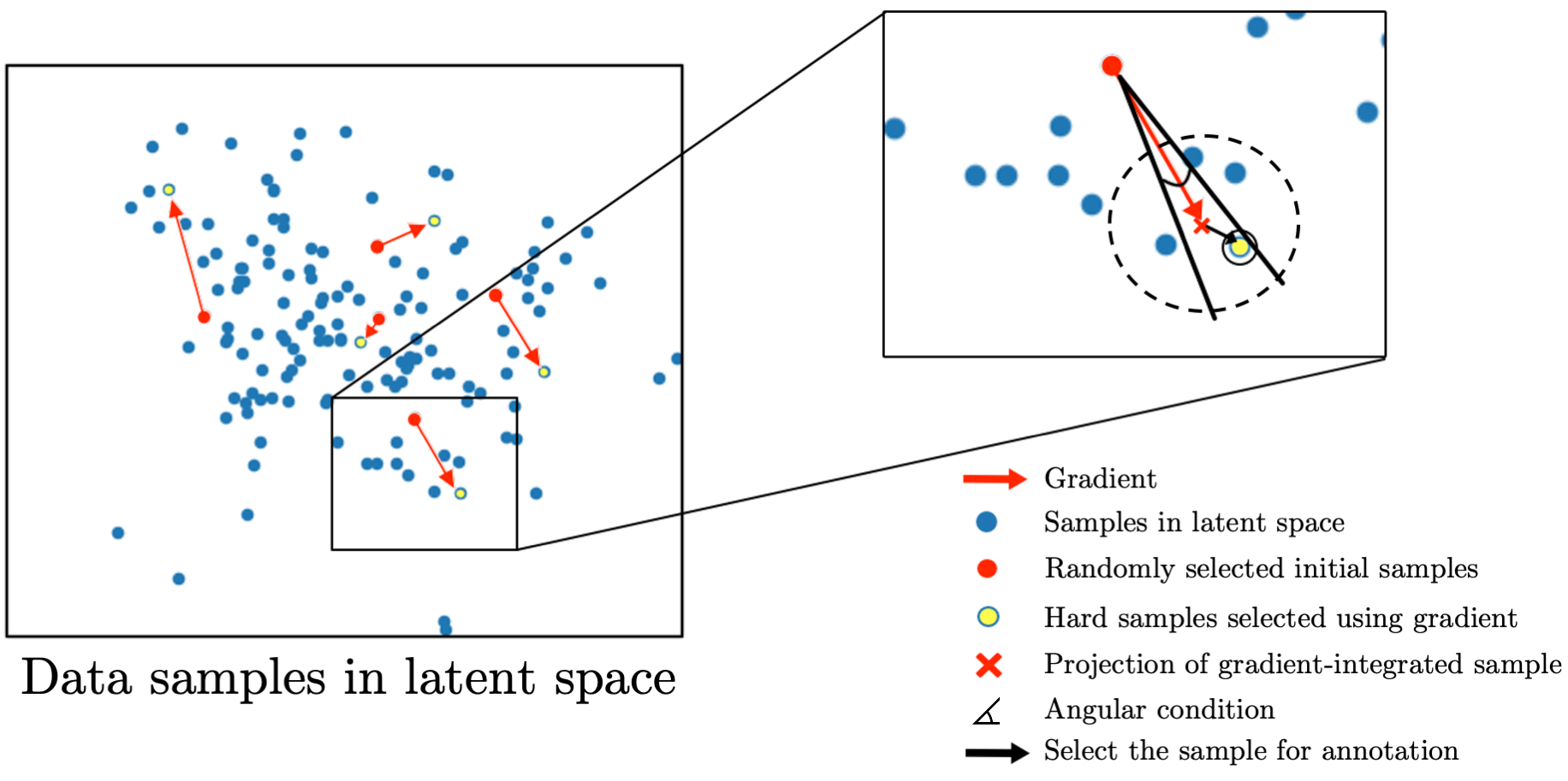}}
\caption{Sampling process in the latent space. The zoomed view shows an example of a new sample suggested by the proposed gradient-guided sampling method.} \label{latent}
\end{figure}

Fig. \ref{latent} illustrates the sampling process in the latent space. The red dots are the latent representation of \(S\) that are randomly selected initially. After training the base model with these samples, for any \(x_{i}\) from \(S\), the loss gradient is backpropagated and integrated to \(x_{i}\) (Eq. \ref{eq:integration}), then projected to \(z'_{i}\) in the latent space (Eq. \ref{eq:projection} and red cross in Fig. \ref{latent}). One of two criteria we use to find an existing real image is to find a \(z_{j}\in Z\) (yellow dot) that has the shortest Euclidean distance to \(z'_{i}\). However, we have found that using the Euclidean distance alone sometimes may fail to find an existing image that is similar to the synthesised image. To mitigate this issue, we introduce an angular condition to limit the search angle (black angle in Fig. \ref{latent}). The angular condition is similar to the cosine distance widely used in machine learning, which in our case was used to constrain the search in the VAE manifold with respect to similar cases.

By using the gradient-guided sampling method, one or more samples can be found given each \(z'_{i}\). To simplify the process, we only select one sample \(z_{j}\in Z\) for each \(z'_{i}\) in our work, which corresponds to image \(x_{j}\in X\). In this way, we select $m$ informative samples and suggest them to the expert for manual annotation. A new training dataset with $m$ more samples \(\hookD'=\{(x_{1},y_{1}),(x_{2},y_{2}),...,(x_{2m},y_{2m})\}\) can be constructed, which consists of initial samples and new samples suggested by the proposed method. The new training set can be used for further training the base network.

\section{Experiments}

\subsection{Dataset}
To demonstrate the proposed sampling method in this paper, we used a brain tumour dataset from the 2019 BraTS Challenge \cite{bakas2018identifying, bakas2017segmentation, bakas2017segmentation2, bakas2017advancing, menze2014multimodal}. The dataset contains T1, T1 gadolinium (Gd)-enhancing, T2 and T2-FLAIR brain MRI volumes for 335 patients diagnosed with high grade gliomas (HGG) or low grade gliomas (LGG), acquired with different clinical protocols from multiple institutions. Two main contributors of the dataset are Center for Biomedical Image Computing and Analytics (CBICA) for 129 patients and The Cancer Imaging Archive (TCIA) for 167 patients. The dataset was pre-processed with skull-striping, interpolation to a uniform isotropic resolution of \(1 mm^{3}\) and registered to SRI24 space \cite{rohlfing2010sri24} with a dimension of $240\times240\times155$. We further processed the provided image volume with zero padding and z-score intensity normalisation. The first and last few image slices of the volume were discarded which were normally blank, resulting in a volume of dimension of $256\times256\times150$ after pre-processing. The annotations of the dataset have four labels: background, Gd-enhancing tumour, the peritumoral edema and the necrotic and non-enhancing tumour core. The latter three labels were combined to form the whole tumour label, which was an important structure for evaluating segmentation accuracy \cite{bakas2018identifying}. The dataset was split into 260/75 for training and test.

\subsection{Experimental design}
The VAE with the encoder and decoder built by residual blocks \cite{szegedy2017inception} was used for learning the data manifold. It was trained for 50 epochs on the images with the Adam optimizer, the loss given by Eq. \ref{eq:vaeloss} and a learning rate of 1e-4. The VAE were trained with empirically selected z dimension numbers of 5. The UNet, as the base model for segmentation, was trained with Adam optimizer using Dice as the loss function and a learning rate of 1e-3. We used a 2D network for segmentation due to the computational constraint, as it is challenging to feed a whole 3D volume to 3D VAE network and segmentation network on a standard GPU that we use (Titan X, 12GB RAM). 

To further reduce annotation cost, we investigate two methods of suggestive annotation: image-wise suggestion and patient-wise suggestion. Ideally image-wise suggestion is preferred because it can select several representative image slices for each patient and the expert does not need to annotate all the slices for the same patient, which may contain high data redundancy. Patient-wise suggestion, on the other hand, would require similar slices from the same patient to be annotated. In reality, however, it may be difficult for the expert to annotate the brain tumour on just a few slices from each patient without assessing other slices in the context due to the reason that the vessel may look similar to a tumour on MR images \cite{bakas2018identifying}. Here we test the two suggestion strategies under an empirical assumption that if one image slice is suggested for annotation by image-wise suggestion method, two adjacent slices would also be given to the expert for reference.

\textbf{Patient-wise suggestion}: Assuming that we use 20 patients' volumes to test our method, we first randomly select 10 patients (equivalent to 150$\times$10 slices) for initial model training of 30 epochs. Then we apply the proposed method to select 10 more volumes from the rest of dataset for suggestive annotation. The model is then trained on the new set of 20 patients for 30 more epochs. To sample the patient in latent space, the latent representations of 150 image slices for each patient was calculated and averaged to form a patient-wise representation.

\textbf{Image-wise suggestion}: To compare with the case of using 20 patients' volumes, 1000 (150*20/3) image slices in total will be used, of which 500 slices will be randomly selected initially to train the model for 30 epochs, then 500 more will be selected by the proposed method. The step size \(\alpha\) is set to 1e-4 for both image- and patient-wise suggestion.

The baseline method is random suggestion, which randomly selects the same number of patients or images for training the segmentation model. An oracle method was also evaluated, which assumes that the annotations are already available for all the data and we select the posteriori `best' (in our case, the most challenging in terms of Dice score) samples according to the goal of optimizing the segmentation model. The oracle method serves as an upper-bound for suggestive annotation. Each experiment was repeated 10 times for evaluating the averaged performance.

We evaluate the performance of our method in two scenarios, \textit{training from scratch} and \textit{transfer learning}. For training from scratch, we sample from the full BraTS dataset. For transfer learning, we use the CBICA dataset to pre-train the UNet for segmentation and then sample from the TCIA dataset for fine-tuning the network with the proposed method. This is common in medical imaging applications, where we have trained a network using data from one site and then need to deploy it to data from another site. In this scenario, TCIA dataset was split into 130/37 for training and test. The data manifold was learnt from TCIA training set only to avoid a data manifold that has bias towards the pre-training (CBICA) dataset. 

\begin{figure}[ht]
\makebox[\textwidth][c]{\includegraphics[width=0.85\textwidth]{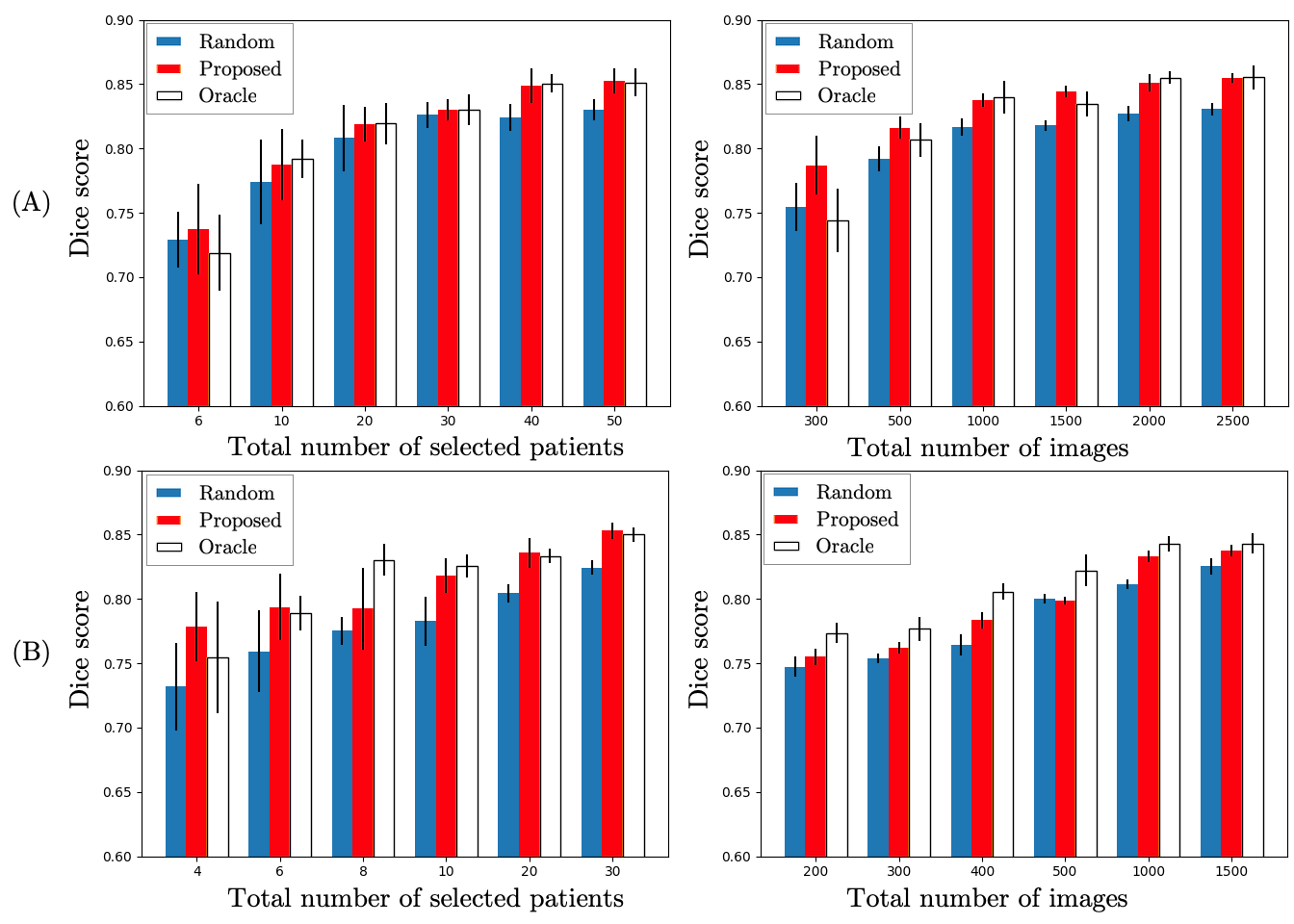}}
\caption{Comparison of the proposed suggestive annotation method with the baseline random suggestion method and the oracle method. Row (A): training from scratch; Row (B): transfer learning from CBICA to TCIA. Left column: patient-wise suggestion; right column: image-wise suggestion. For the total number of $n$ patients or images, $n/2$ are randomly selected for initial model training and $n/2$ are selected using one of the methods: random, proposed or oracle.} \label{result}
\end{figure}

\begin{figure}[htp]
\includegraphics[width=0.8\textwidth]{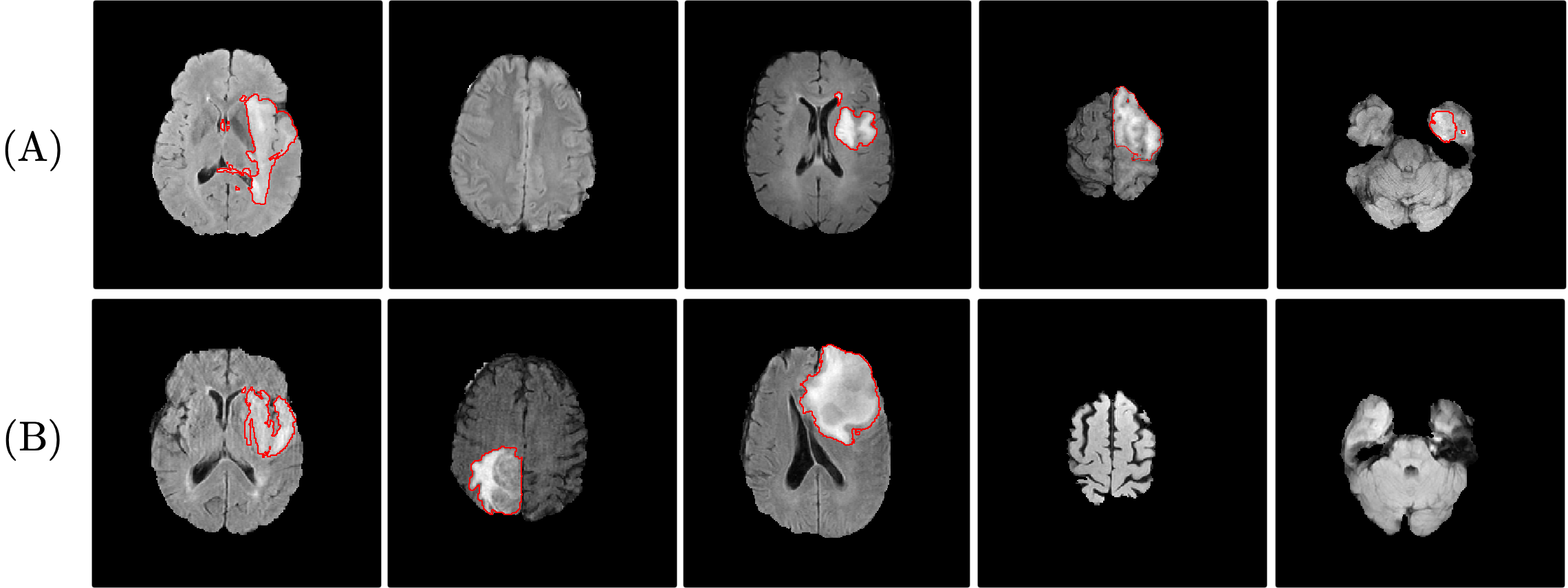}
\caption{Interpretation of the proposed method. Row (A): initial samples (red contour: whole tumour), which provide the gradient for search; Row (B): samples suggested by proposed method, which appear quite different from the initial samples.}
\label{fig:qual}
\end{figure}

\subsection{Results}
Fig. \ref{result} compares the performance of the proposed method to the baseline random suggestion method and the oracle method, in terms of the Dice score for segmentation. It shows that when training from scratch (Fig. \ref{result}A), the proposed method outperforms the random suggestion under all circumstances. In addition, it yields similar or sometimes better performance compared to the oracle method. Training with 50 out of the full 260 volumes in BraTS dataset (19\%) or 2,500 out of 39,000 image slices (7\%) suggested by the proposed method achieved comparable performance to training on full dataset (a Dice score of 0.853 on full dataset is observed). For the transfer learning scenario (Fig. \ref{result} (B)), the proposed method also consistently outperformed random suggestion and achieved comparable performance as the oracle method. Some examples of images suggested by proposed method for annotation are given in Fig. \ref{fig:qual}.

When we compare patient-wise suggestion to image-wise suggestion, the latter is better in annotation efficiency since much less images slices are suggested for annotation. In Fig. \ref{result} (A), 2,500 images are suggested for annotation, which can achieve a similar performance as annotating 50 volumes (7,500 images). Even though some image slices are needed to provide the expert with the context in image-wise suggestion, given the expert would not need to annotate these context images, image-wise suggestion is still a more efficient method.

\section{Conclusions}
In this paper, we present a gradient-guided suggestive annotation framework and demonstrate it for brain tumour image segmentation. The experimental results show that selecting informative samples substantially reduces the annotation costs and benefits model training. The proposed method achieved a similar performance as the oracle method. Moreover, with only 19\% patient volumes or 7\% image slices suggested by our method, the segmentation model can achieve a comparable performance to training on the full dataset. The proposed method is easy to generalise to other segmentation tasks and it has a great potential to lower the annotation cost in medical imaging applications.

\section{Acknowledgements}
This research is independent research funded by the NIHR Imperial Biomedical Research Centre (BRC). The views expressed in this publication are those of the author(s) and not necessarily those of the NHS, NIHR or Department of Health. We gratefully acknowledge the support of NVIDIA Corporation with the donation of the GPU used for this research.

\bibliographystyle{splncs}
\bibliography{ref}
\end{document}